\documentclass[10pt,twocolumn,letterpaper]{article}

\usepackage{3dv}
\usepackage{times}
\usepackage{epsfig}
\usepackage{graphicx}
\usepackage{amsmath}
\usepackage{amssymb}

\usepackage{amsmath}
\usepackage{amssymb}
\usepackage{xcolor}
\usepackage{subcaption}
\usepackage{tabularx}
\usepackage{stmaryrd}
\usepackage{mathrsfs}
\usepackage{url}

\definecolor{TabRowColor}{RGB}{240, 240, 240}

\definecolor{ColorGray}{RGB}{96, 96, 96}
\definecolor{ColorDarkGreen}{RGB}{19, 179, 50}
\definecolor{ColorLightBlue}{RGB}{18, 137, 255}
\definecolor{ColorOrange}{RGB}{252, 111, 3}

\newcommand{\myparagraphcpt}[1]{\noindent\textbf{#1}~}
\newcommand{\myparagraph}[1]{\vspace{2.6mm}\noindent\textbf{#1}~}

\newcommand{\tbf}[1]{\textbf{#1}}
\newcommand{\tdf}[1]{{\color{gray}#1}}
\newcommand{\tul}[1]{\underline{#1}}

\def\OurMethodName{SRFeat}

\usepackage[pagebackref=true,breaklinks=true,colorlinks,bookmarks=false]{hyperref}

\usepackage[capitalize]{cleveref}
\crefname{section}{Sec.}{Secs.}
\Crefname{section}{Section}{Sections}
\Crefname{table}{Table}{Tables}
\crefname{table}{Tab.}{Tabs.}

\threedvfinalcopy % *** Uncomment this line for the final submission

\def\threedvPaperID{***} % *** Enter the 3DV Paper ID here

\begin{document}

%%%%%%%%% TITLE
\title{\OurMethodName{}: Learning Locally Accurate and Globally Consistent\\Non-Rigid Shape Correspondence}

\author{Lei Li \hspace{1cm} Souhaib Attaiki \hspace{1cm} Maks Ovsjanikov\\
LIX, \'{E}cole Polytechnique, IP Paris
}

\maketitle
%\thispagestyle{empty}

%%%%%%%%% ABSTRACT
\begin{abstract}
   In this work, we present a novel learning-based framework that combines the local accuracy of contrastive learning with the global consistency of geometric approaches, for robust non-rigid matching. We first observe that while contrastive learning can lead to powerful point-wise features, the learned correspondences commonly lack smoothness and consistency, owing to the purely combinatorial nature of the standard contrastive losses. To overcome this limitation we propose to boost contrastive feature learning with two types of smoothness regularization that inject geometric information into correspondence learning. With this novel combination in hand, the resulting features are both highly discriminative across individual points, and, at the same time, lead to robust and consistent correspondences, through simple proximity queries. Our framework is  general and is applicable to local feature learning in both the 3D and 2D domains. We demonstrate the superiority of our approach through extensive experiments on a wide range of challenging matching benchmarks, including 3D non-rigid shape correspondence and 2D image keypoint matching.
\end{abstract}

%%%%%%%%% BODY TEXT
\section{Introduction}
\label{sec:Introduction}

Finding accurate correspondences across geometric objects is a fundamental task in a wide range of computer vision and graphics problems, such as object tracking, registration, texture transfer,  and statistical shape analysis~\cite{zhou2016fast,dinh2005texture,bogo2014faust}, among many others.
The presence of significant variations in 3D or 2D geometric objects, including rigid and non-rigid transformations, makes it challenging to develop a single unified theoretical deformation model for robust matching~\cite{van2011survey,tam2012registration}.
Earlier approaches to computing correspondences heavily relied on hand-crafted features and pipelines~\cite{van2011survey}.
In more recent years, there has been a growing body of literature advocating the use of deeply \emph{learned} features that demonstrate superior matching performance over axiomatic approaches~\cite{FeyDGMC,Sarlin_2020_CVPR,rostami2019survey,Choy_2019_ICCV,Donati_2020_CVPR}.

\begin{figure}[t!]
   \centering
   \includegraphics[width=\linewidth]{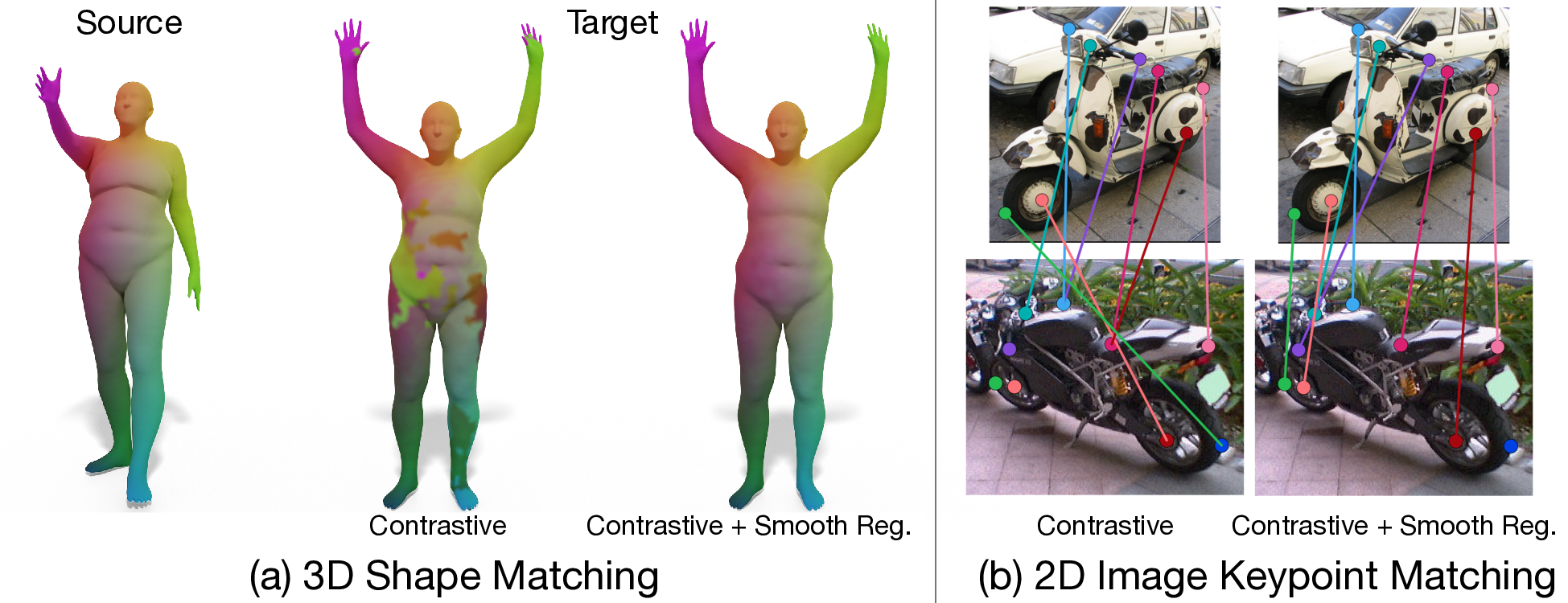}
   \caption{We propose a smoothness-regularized contrastive learning approach for local feature learning, which leads to more accurate, \emph{consistent} correspondences. (a) Dense non-rigid 3D shape matching via learned features on the FAUST dataset~\cite{bogo2014faust}. Correspondence is visualized by color transfer. (b) 2D image keypoint matching through node-wise features on the PASCAL VOC dataset~\cite{everingham2010pascal}. Ground-truth corresponding keypoints have the same node color.\vspace{-2mm}}
  \label{fig:teaser}
\end{figure}

In this work, we focus on learning discriminative \emph{local features} that can robustly identify each point on a given geometric object for correspondence.
Given such local features for a pair of objects, finding point-wise correspondences simply reduces to proximity queries in the learned feature space~\cite{Guo:2016:CPELFD}.
However, it is not easy to endow local features with descriptiveness and robustness for geometric objects potentially undergoing arbitrary deformations.

Contrastive learning is a popular approach to training local feature extractors, for example, in the task of 3D rigid point cloud registration~\cite{Zeng_2017_CVPR,Deng_2018_CVPR,Gojcic_2019_CVPR,Li_2020_CVPR,Bai_2020_CVPR,Choy_2019_ICCV,xie2020pointcontrast}.
The wide adoption of contrastive learning lies in the fact that it is extremely generic, as also actively studied in 2D visual representation learning~\cite{he2020momentum,chen2020simple,grill2020bootstrap,FeyDGMC}, and thus can be applicable to arbitrary 3D or 2D shape classes.
On the other hand, this learning paradigm is inherently based on computing correspondences across \emph{individual points} by comparing their local features.
As a result, while the learned features can be very discriminative, the overall quality of the correspondences can suffer and especially lack smoothness and consistency~\cite{choy2020deep,bai2021pointdsc}, as shown in \cref{fig:teaser} (a).
Thus, so far, there is limited success in training local features purely with contrastive learning for direct nearest-neighbor matching, e.g., for non-rigid shape correspondence~\cite{deng2022survey}.

Motivated by the above discussion, we introduce a novel smoothness-regularized contrastive learning approach, enabling robust feature-based matching of deformable objects.
Specifically, we boost the contrastive loss at training time (\eg, the PointInfoNCE loss~\cite{xie2020pointcontrast}) with  powerful smoothness regularization terms that promote the \emph{overall consistency} of the learned point-wise features.
Our resulting approach, which we call \emph{\OurMethodName{}}, enjoys the advantages of contrastive learning, by obtaining highly discriminative local features, which can be accurately matched via direct feature proximity queries.
Moreover, owing to the smoothness promotion, the resulting local features are strongly regularized, thus leading to overall smooth and consistent correspondences even without any post-processing (\cref{fig:teaser} (a)).

We initiate the study of smoothness regularization for contrastive learning, specifically, for deformable shape correspondence.
We propose two implementation variants for the smoothness regularization at training time (\cref{subsec:SmoothnessRegularization}): (1) a Dirichlet energy loss that penalizes discontinuities in the feature space and (2) a spectral loss that evaluates the correspondence matrices in the spectral domain.
We demonstrate the superior performance of \OurMethodName{} through a comprehensive set of experiments on diverse non-rigid shape matching benchmarks~\cite{bogo2014faust,anguelov2005scape,MelziMRCRPWO19,Zuffi_2017_CVPR}.
At test time, we compute correspondences between non-rigid shapes by nearest-neighbor queries with the learned features, in contrast to the state-of-the-art methods~\cite{Donati_2020_CVPR,eisenberger2020deep} that typically require the Laplacian basis computation and a test-time correspondence optimization in the spectral domain.

In addition, our smoothness regularization is remarkably generic and can be easily incorporated into other modern contrastive feature learning frameworks in other domains (\eg, images),
and thus we position \OurMethodName{} as a general local feature learning approach.
To demonstrate the wide applicability, as shown in \cref{fig:teaser} (b), we apply \OurMethodName{} to the 2D image domain for keypoint matching~\cite{everingham2010pascal,FeyDGMC}, bringing significant improvement over existing methods.

In a nutshell, the main contributions of our work are as follows:
(1) We introduce a novel generic smoothness regularization to the contrastive feature learning framework, substantially improving the smoothness and consistency of correspondences found by local features;
(2) We establish a link between contrastive learning and spectral (functional map-based) shape correspondence methods by relating ways in which these approaches operate on the computed correspondence matrices;
(3) We show that contrastive feature learning combined with smoothness regularization yields superior matching performance over existing methods on widely adopted non-rigid shape benchmarks;
(4) We demonstrate the strong generality of \OurMethodName{} to the 2D domain for tackling matching problems on real-world image data.
Our code and data are publicly available\footnote{\url{https://github.com/craigleili/SRFeat}}.

\section{Related Work}
\label{sec:RelatedWork}

\myparagraphcpt{Contrastive Learning}
Contrastive learning has recently received significant research attention as a powerful representation learning paradigm for both 2D and 3D data.
In the 2D domain, contrastive learning is widely used for unsupervised learning of 2D representations~\cite{oord2018representation,he2020momentum,chen2020simple,grill2020bootstrap,FeyDGMC,Cui_2021_ICCV}.
Meanwhile, researchers also actively investigate this generic learning paradigm for 3D geometric data.
For example, PointContrast~\cite{xie2020pointcontrast} and its follow-up work~\cite{Hou_2021_CVPR} perform local feature contrasting at the point level between two transformed 3D scene fragments.
The learned feature representation is shown to be useful in downstream 3D tasks like segmentation and detection~\cite{dai2017scannet}.
Besides, there also exist recent works performing the contrasting at the local patch level~\cite{du2021self}, object instance level~\cite{rao2021randomrooms}, global scene level~\cite{huang2021spatio}, or both the shape and point levels~\cite{wang2021unsupervised}.
Our work, related to~\cite{xie2020pointcontrast,FeyDGMC}, focuses on
learning \textit{local features} for robustly identifying {individual points} on deformable geometric objects for correspondence.
To boost the performance of contrastive feature learning, we propose smoothness regularization and show its utility in a wide range of matching problems.

\myparagraph{Shape Matching}
Shape matching is a key problem in 3D shape analysis and has been extensively studied in recent decades~\cite{Guo:2016:CPELFD,van2011survey,biasotti2016recent,guo2020deep,bronstein2017geometric,rostami2019survey,deng2022survey}.
Earlier works focused primarily on hand-crafted 3D local features for matching, including both extrinsic~\cite{Johnson:1999:SPIN,Frome:2004:3DShapeContext,Rusu:2008:PFH,Rusu:2009:FPFH,Salti:2014:SHOTUS} and intrinsic~\cite{sun2009concise,aubry2011wave} descriptors.
In recent years, research focus has shifted to learned local features for better robustness in matching, for example, in the task of point cloud registration~\cite{Zeng_2017_CVPR,Deng_2018_CVPR,Gojcic_2019_CVPR,Wang_2019_ICCV,Yew_2020_CVPR,Li_2020_CVPR,ao2020spinnet,Bai_2020_CVPR,bai2021pointdsc}.
For non-rigid shapes, a common approach to computing correspondence is to leverage spectral information, e.g., using the functional map framework~\cite{ovsjanikov2012functional} and its follow-up works~\cite{kovnatsky2013coupled,huang2014functional,burghard2017embedding,rodola2017partial,nogneng2017informative,ren2018continuous,huang2017adjoint}. In particular,
several recent approaches~\cite{Litany_2017_ICCV,roufosse2019unsupervised,eisenberger2020deep,Groueix_2018_ECCV,Donati_2020_CVPR,attaiki2021dpfm} have built upon this framework by advocating learned probe functions.
There also exist a few works~\cite{Masci_2015_ICCV_Workshops,Mitchel_2021_ICCV} exploring surface CNNs with the contrastive loss proposed by Hadsell \etal \cite{hadsell2006dimensionality} to learn features for non-rigid shapes. However, the performance of such approaches on dense shape correspondence was not shown to be comparable to that of the spectral methods.

Our \OurMethodName{} framework differs from the above state-of-the-art non-rigid shape matching approaches~\cite{Donati_2020_CVPR,attaiki2021dpfm}, which require the spectral basis computation and optimization in the spectral domain at test time.
Instead, our approach achieves superior matching performance by directly matching local features, learned via our smoothness-regularized contrastive learning strategy.

\myparagraph{Image Keypoint Matching}
Image matching is a well studied area in computed vision, and a full review is beyond the scope of this work. We refer the interested readers to recent surveys \cite{Ma2020,Lenc2018LargeSE,Balntas2017HPatchesAB} for a more in-depth discussion. 
Finding correspondences between 2D images is a difficult problem, due to the potentially strong differences in appearance, and ambiguities introduced by repeating patterns. Classical methods for solving this problem were based on handcrafted features such as \cite{Lowe2004,Mikolajczyk2002,Tuytelaars2007}. Strategies such as ratio test \cite{Lowe2004} or mutual check were used to reduce ambiguous matches. Recent methods are based on trainable feature descriptors extracted by convolutional neural networks (CNNs). They either operate on patches extracted by handcrafted feature detectors and produce a sparse set of descriptors \cite{Simonyan2014,BMVC2016_119,Balntas2016PNNetCT,SimoSerra2015}, or involve end-to-end methods combining detection and description \cite{Yi2016,Noh2017LargeScaleIR,Choy2016UniversalCN}. 
Several methods have been proposed for producing consistent matches~\cite{Rocco18b,Zanfir_2018_CVPR,wang2019learning,FeyDGMC}.
DGMC~\cite{FeyDGMC} tackles this problem by constructing a graph for image keypoints and seeking consensus of matches in local neighborhoods using a synchronous message passing network, while using a standard contrastive loss for training.
In this work, 
we investigate the strong generality of our smoothness regularization in the image keypoint matching task, showing its significant improvement to the contrastive learning used in~\cite{FeyDGMC}.

\section{Method}
\label{sec:Method}

\subsection{Background}
\label{subsec:Background}

Contrastive learning~\cite{hadsell2006dimensionality,oord2018representation} is a widely adopted approach to learning informative representations for 3D~\cite{xie2020pointcontrast} and 2D~\cite{he2020momentum,FeyDGMC} vision understanding tasks.
Specifically, the PointInfoNCE loss introduced in~\cite{xie2020pointcontrast} was formulated on individual points to train 3D local features for rigid alignment.
Given a pair of point clouds $\mathcal{P}_1$ and $\mathcal{P}_2$ with $n_1$ and $n_2$ points, respectively, below we re-write this contrastive loss using \emph{a feature similarity matrix} $\mathbf{\Pi} \in \mathbb{R}^{n_1 \times n_2}$, as this will be useful to establish the link to spectral approaches in \cref{subsec:SmoothnessRegularization}.
Let $\mathbf{f}_{1}^{i}, \ \mathbf{f}_{2}^{j} \in \mathbb{R}^{d}$ denote $d$-dimensional features for the $i^{\text{th}}$ point in $\mathcal{P}_1$ and $j^{\text{th}}$ point in $\mathcal{P}_2$, respectively.
The similarity matrix $\mathbf{\Pi}$ is then constructed as:
	\vspace{-0.5mm}
\begin{align}
	& \mathbf{\Pi}^{i,j} = \dfrac{\exp\big(s(\mathbf{f}_{1}^{i}, \mathbf{f}_{2}^{j}) / \tau\big)}{\sum_{k=1}^{n_2}\exp\big(s(\mathbf{f}_{1}^{i}, \mathbf{f}_{2}^{k}) / \tau\big)},
	\label{eq:SoftPointCorrMatrix}\\
	&s(\mathbf{x}, \mathbf{y}) = \mathbf{x} \cdot \mathbf{y},
	\label{eq:FeatureDistanceFunc}
\end{align}
where $s(\cdot,\cdot)$ is the similarity measurement between two feature vectors, and $\tau$ is a temperature hyper-parameter.
\cref{eq:SoftPointCorrMatrix} can be thought of as applying row-wise softmax to the pairwise similarity in \cref{eq:FeatureDistanceFunc}.
The PointInfoNCE loss is then defined as:
\begin{equation}
	\mathcal{L}_{\text{c}} = -\sum_{i} \log\big(\mathbf{\Pi}^{i, \text{GT}(i)}\big).
	\label{eq:ContrastiveLoss}
\end{equation}
Here $\text{GT}(i)$ denotes the ground-truth correspondence.
In~\cite{xie2020pointcontrast}, $\mathcal{P}_1$ and $\mathcal{P}_2$ are partially overlapped and thus only points sparsely sampled in the overlap region are considered in the above formulation.

Another example of contrastive learning is its application to the graph matching problem~\cite{wang2019learning}, aiming at establishing correspondences between the nodes of two input graphs, such as for 2D image keypoint matching~\cite{everingham2010pascal}.
The training loss used in the deep graph matching consensus (DGMC) framework~\cite{FeyDGMC} has a similar formulation to \cref{eq:ContrastiveLoss}.

Interestingly, as we demonstrate in \cref{subsec:3DShapeMatching}, the PointInfoNCE loss alone, when used in conjunction with a recent powerful feature extractor \cite{Sharp2020DiffusionNetDA}, can already lead to point-wise features that induce competitive  correspondences on near-isometric 3D shape benchmarks.
This result suggests that it is viable to establish correspondences between non-rigid shape pairs by simple proximity queries in the feature space, differently from common approaches~\cite{boscaini2016learning,wiersma2020cnns,Sharp2020DiffusionNetDA,Donati_2020_CVPR} that are either based on predicting vertex ids of some reference shape or use test time optimization. However, we also find that using the PointInfoNCE loss alone can lead to \textit{discontinuous} maps, especially in the presence of non-isometries, due to the issues discussed below.

\myparagraph{Formulation Issues}
Despite the generality of contrastive learning, fundamentally the loss in~\cref{eq:ContrastiveLoss} only considers whether individual point correspondences are correct or not, without exploiting any geometric structure of the shapes.
This means that incorrect predictions are penalized equally regardless of whether they are close to the ground-truth match or not.
More broadly, contrastive learning does not consider structural properties of the underlying map, such as continuity or smoothness. Such properties emerge when one considers either \emph{relations} between learned feature embeddings of points on the shapes, or analyzing the  map as a whole~\cite{Litany_2017_ICCV,roufosse2019unsupervised,Donati_2020_CVPR}.
To mitigate the issues, we propose to use simple yet powerful regularization by promoting smoothness in the learned local features, while maintaining the simplicity and advantages of contrastive learning.

\subsection{Smoothness-Regularized Feature Learning}
\label{subsec:SmoothnessRegularization}

Our goal is to train a robust feature extraction network that would enable shape matching via simple proximity queries between features at different points. 
We base our approach, \OurMethodName{}, on the generic contrastive learning framework (\cref{subsec:Background}), and propose to boost contrastive feature learning with smoothness regularization.
In what follows, we first instantiate the formulations of smoothness regularization in the 3D domain and then discuss the generalization to the 2D domain (\cref{subsec:GeneralizationTo2DMatching}).

Let us first define the notation for feature extraction.
Suppose we are given a pair of non-rigid shapes $\mathcal{S}_1$ and $\mathcal{S}_2$ with ground-truth correspondences.
The shapes are represented as graphs (\ie, triangle meshes) and contain $n_1$ and $n_2$ vertices, respectively.
We denote our feature extractor as $\mathcal{F}_\Theta$, where $\Theta$ represents the trainable network parameters.
We obtain point-wise features for each shape by forwarding it to the feature extractor $\mathcal{F}_\Theta$.
Specifically, let $\mathbf{G}_1 = \mathcal{F}_\Theta(\mathcal{S}_1)$ and $\mathbf{G}_2 = \mathcal{F}_\Theta(\mathcal{S}_2)$ denote the sets of resulting $d$-dimensional features for the two shapes, respectively, where $\mathbf{G}_1 \in \mathbb{R}^{n_1 \times d}$ and $\mathbf{G}_2 \in \mathbb{R}^{n_2 \times d}$.
We refer to each column of the feature matrices as a real-valued feature function defined on the vertices of the corresponding shape.

\begin{figure}[t!]
   \centering
   \includegraphics[width=\linewidth]{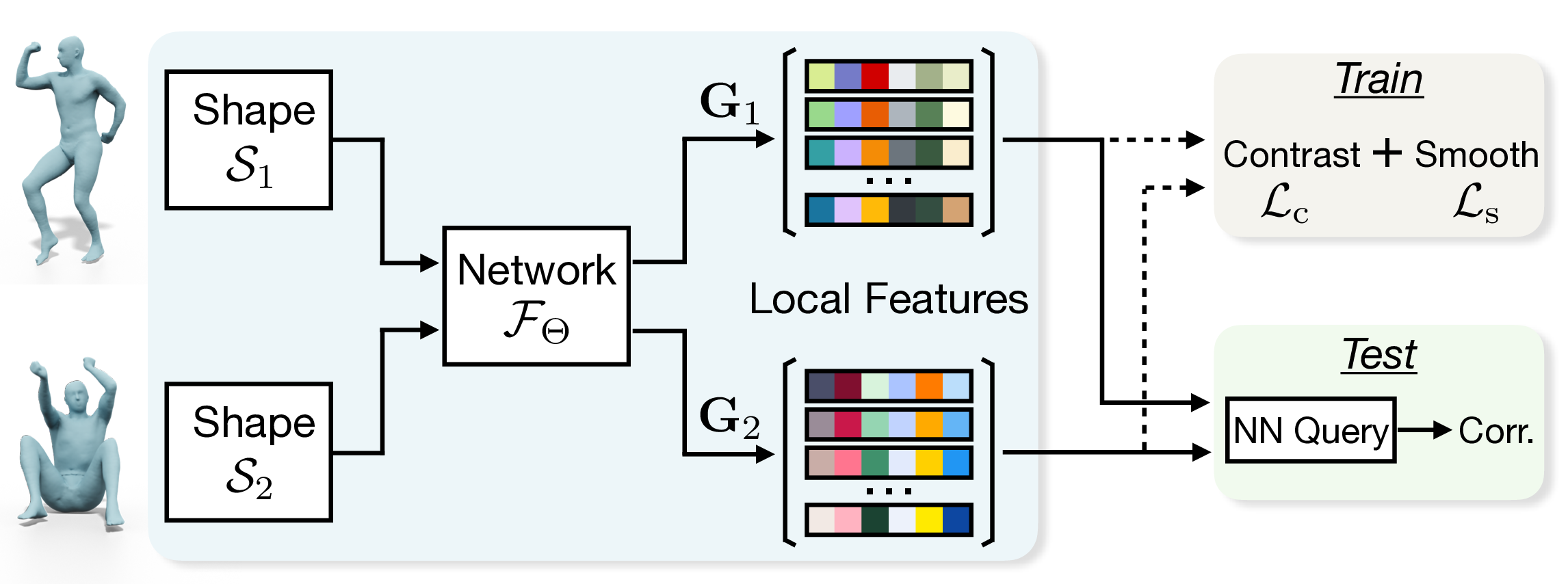}
   \caption{Overview of our smoothness-regularized contrastive learning framework. To train the extracted local features $\mathbf{G}_1$ and $\mathbf{G}_2$ by the network $\mathcal{F}_{\Theta}$, we augment the contrastive loss $\mathcal{L}_{\text{c}}$ with a smoothness regularization loss $\mathcal{L}_{\text{s}}$. At test time, correspondences between the input deformable shapes are found by nearest-neighbor query between $\mathbf{G}_1$ and $\mathbf{G}_2$.}
  \label{fig:pipeline}
\end{figure}

Our \OurMethodName{} framework is conceptually simple and is illustrated in \cref{fig:pipeline}.
\OurMethodName{} trains the feature extractor $\mathcal{F}_\Theta$ with a novel training loss $\mathcal{L}$ that combines contrastive learning with smoothness regularization:
\begin{equation}
	\label{eq:OurLoss}
	\mathcal{L} = \mathcal{L}_{\text{c}} + \lambda \mathcal{L}_{\text{s}},
\end{equation}
where $\lambda$ is a weighting hyper-parameter.
The first term $\mathcal{L}_{\text{c}}$ is the contrastive loss (\cref{subsec:Background}) that promotes small feature distance between corresponding points and large feature distance between non-corresponding ones.
The second term $\mathcal{L}_{\text{s}}$ is the smoothness regularization loss for $\mathbf{G}_1$ and $\mathbf{G}_2$, exploiting geometric information in the input shapes.
We propose two variants for $\mathcal{L}_{\text{s}}$, as detailed below.

\myparagraph{Dirichlet Energy Loss}
To regularize the smoothness of the learned local features, we consider the \textit{feature functions} defined on the shapes (\ie, each column of $\mathbf{G}_1$ and $\mathbf{G}_2$).
Our first loss is based on the Dirichlet energy~\cite{pinkall1993computing}, which intuitively measures how smooth those functions are.
Given a real-valued function $g: \mathcal{S} \rightarrow \mathbb{R}$ on the shape $\mathcal{S}$, the Dirichlet energy is defined as:
\begin{equation}
	\label{eq:DirichletEnergySmooth}
	\mathcal{E}_{\text{d}}(g) = \int_{ \mathcal{S}} \| \nabla g \|^2 dA.
\end{equation}
In the discrete case, the Dirichlet energy can be computed as:
\begin{equation}
	\label{eq:DirichletEnergyDiscrete}
	\mathcal{E}_{\text{d}}(\mathbf{g}) = \mathbf{g}^{\top} \mathbf{W} \mathbf{g},
\end{equation}
where $\mathbf{g}$ is a vector representing the input function, and $\mathbf{W}$ denotes the classical symmetric cotangent weight (stiffness) matrix~\cite{pinkall1993computing}.

We then formulate our Dirichlet energy loss as:
\begin{equation}
	\label{eq:DirichletEnergyLoss}
	\mathcal{L}_{\text{s}} = \frac{1}{2d} \sum_{i} \mathcal{E}_{\text{d}}(\mathbf{G}_1^{i}) + \frac{1}{2d} \sum_{i} \mathcal{E}_{\text{d}}(\mathbf{G}_2^{i}),
\end{equation}
where $\mathbf{G}_1^{i}$ denotes the $i$-th column of $\mathbf{G}_1$, and similarly for $\mathbf{G}_2^{i}$.
Intuitively, our Dirichlet energy loss promotes structural properties of the feature functions by considering the columns of the feature matrices $\mathbf{G}_1, \mathbf{G}_2$, while the contrastive loss $\mathcal{L}_{\text{c}}$ in \cref{eq:ContrastiveLoss} supervises the per-point features, \ie, \textit{the rows} of  $\mathbf{G}_1, \mathbf{G}_2$.
Thus combining both losses enables the network to produce discriminative and globally consistent features, leading to more accurate, smooth correspondences.

\myparagraph{Spectral Loss} 
We propose another variant of smoothness regularization by going into the spectral domain to examine correspondences between input shapes as a whole.
Existing spectral methods for non-rigid shape matching, particularly, the functional map-based ones~\cite{ovsjanikov2012functional,roufosse2019unsupervised,Donati_2020_CVPR,attaiki2021dpfm}, encode correspondences in a reduced spectral basis, resulting in small-sized matrices that come with a suite of theoretical and computational tools and allow to enforce geometric consistency, which otherwise is computationally prohibitive.
On the other hand, using a reduced basis normally leads to the loss of local or high-frequency details in the matching.
Inspired by this line of works, we propose to combine the advantages of global consistency imposed by spectral representations with the precision of a local point-wise contrastive loss for robust feature learning.

Our starting point is that the feature similarity matrix $\mathbf{\Pi}$ defined in \cref{eq:SoftPointCorrMatrix} can be thought of as a \textit{soft} point-wise map, where each row stores the probability distribution for a point in $\mathcal{S}_1$ to be matched to points in $\mathcal{S}_2$. A point-wise map can be interpreted as a functional map in the complete basis, and we exploit the above idea that encoding the map in a reduced spectral basis can introduce global information.
Specifically, given the soft point-wise map $\mathbf{\Pi}$, we compute its associated functional map $\mathbf{C} \in \mathbb{R}^{k \times k}$ by projecting it onto a reduced spectral basis:
\begin{equation}
    \label{eq:SoftPointwiseMapToFunctionalMap}
    \mathbf{C} = \mathbf{\Phi}^{\dagger}_{1} \mathbf{\Pi} \mathbf{\Phi}_{2}, 
\end{equation}
where $\mathbf{\Phi}_{1} \in \mathbb{R}^{n_1 \times k}$ and $\mathbf{\Phi}_{2} \in \mathbb{R}^{n_2 \times k}$ are matrices storing, as columns, the first $k$ eigenfunctions of the Laplace-Beltrami operator~\cite{vallet2008spectral} on the respective shape, and $\dagger$ denotes the Moore-Penrose inverse.
When the eigenfunctions are orthonormal \wrt{} the area-weighted inner product $\mathbf{\Phi}_{1}^{\top} \mathbf{A}_{1} \mathbf{\Phi}_{1} = \mathbf{I}$, then \cref{eq:SoftPointwiseMapToFunctionalMap} can be written as $\mathbf{C} = \mathbf{\Phi}_{1}^{\top} \mathbf{A}_{1} \mathbf{\Pi} \mathbf{\Phi}_{2}$.
Note that $k$ is typically in the range $[20, 100]$ and thus $\mathbf{C}$ is orders of magnitude smaller than $\mathbf{\Pi}$.

At last, we define our spectral loss as:
\begin{equation}
	\label{eq:SpectralFMapLoss}
	\mathcal{L}_{\text{s}} = \| \mathbf{C} - \mathbf{C}_{\text{gt}} \|^2,
\end{equation}
where $\mathbf{C}_{\text{gt}}$ is the ground-truth functional map computed according to \cref{eq:SoftPointwiseMapToFunctionalMap} but with a binary matrix $\mathbf{\Pi}_{\text{gt}}$ representing the ground-truth point-wise map.

\begin{figure}[t!]
	\centering
	\includegraphics[width=\linewidth]{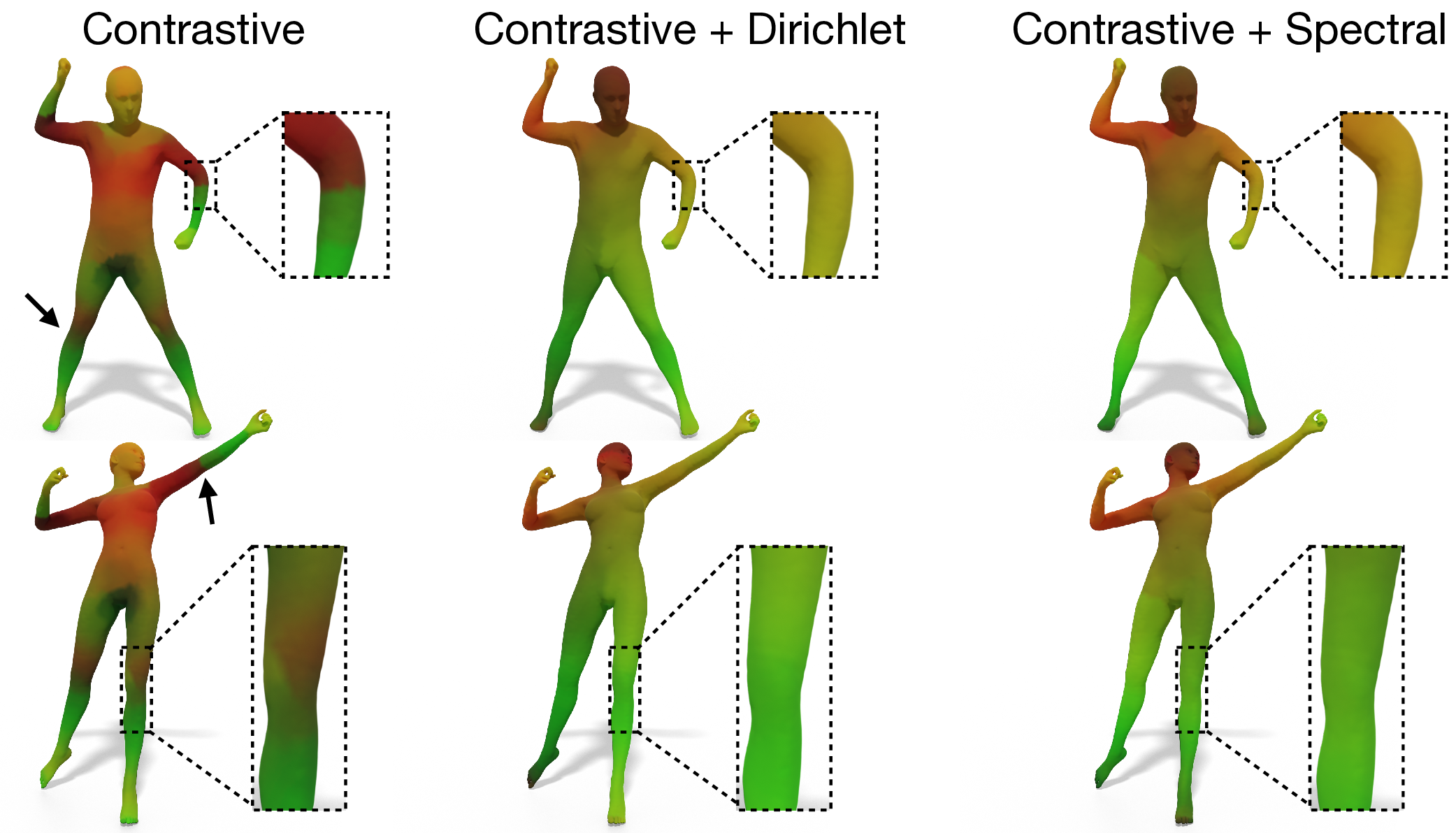}
	\caption{Effect of smoothness regularization on learned local features for a shape pair. 128-dimensional local features are projected to 2D by t-SNE~\cite{van2008visualizing} and encoded as colors.}
	\label{fig:feature_smoothness}
\end{figure}

\myparagraph{Discussion} 
To sum up, our \OurMethodName{} framework introduces consistency into the contrastive feature learning process through either the Dirichlet energy loss \cref{eq:DirichletEnergyLoss}, which regularizes the distribution of features on the points of the domain directly, or the spectral loss \cref{eq:SpectralFMapLoss}, which regularizes correspondences in the reduced spectral basis.
\cref{fig:feature_smoothness} visualizes the effectiveness of smoothness regularization on the learned local features.

We reiterate that \OurMethodName{} differs from the currently dominant functional map-based approaches, like GeomFmaps \cite{Donati_2020_CVPR} and DPFM \cite{attaiki2021dpfm}, for non-rigid shape matching.
Firstly, the above works rely on the Laplacian basis and on solving for the optimal  functional map matrix at test time, whereas our approach uses nearest-neighbor search directly in the local feature domain.
Secondly, since we compute correspondences in the complete basis, foregoing the functional map estimation, our approach can lead to more precise matches and better performance than the above works that only operate in the reduced spectral basis (\cref{sec:Experiments}).
Lastly, in our spectral loss \cref{eq:SoftPointwiseMapToFunctionalMap} we compute $\mathbf{C}$ directly from a learned soft point-wise map, which avoids the need for \textit{solving linear systems} inside the network, in contrast to all previous deep functional maps approaches~\cite{Litany_2017_ICCV,Donati_2020_CVPR,roufosse2019unsupervised,Halimi_2019_CVPR,attaiki2021dpfm}, which require differentiating through the matrix inverse, and which can be numerically unstable during training as observed in~\cite{roufosse2019unsupervised,Donati_2020_CVPR}.

\subsection{Generalization to 2D Matching}
\label{subsec:GeneralizationTo2DMatching}

Our \OurMethodName{} framework naturally inherits the generality of contrastive learning, although our smoothness regularization losses are firstly formulated in the 3D domain.
Indeed, the graph matching problem mentioned in \cref{subsec:Background} can also benefit from smoothness regularization.
Concretely, for the image keypoint matching task~\cite{everingham2010pascal}, each object instance in images is annotated with a set of keypoints for matching.
To apply our smoothness regularization, we need to construct a graph for the keypoints of each image object. For this, we adopt the Delaunay triangulation, similar to~\cite{FeyDGMC}. 
The resulting graphs are 2D meshes, and we compute the Laplacian matrix for them in a similar manner to 3D meshes.
With this simple adaptation, we can compute the smoothness regularization loss $\mathcal{L}_{\text{s}}$, and combine it with the contrastive loss \cref{eq:ContrastiveLoss} to guide the feature learning for graph nodes, as done in \cref{eq:OurLoss}.
More implementation details are provided in \cref{subsec:ImageMatching}.

\section{Experiments}
\label{sec:Experiments}

We evaluate our \OurMethodName{} framework on a wide range of challenging deformable shape matching tasks.
First, in \cref{subsec:3DShapeMatching}, we conduct experiments on existing 3D non-rigid shape correspondence benchmarks, including human shape datasets like FAUST~\cite{bogo2014faust}, SCAPE~\cite{anguelov2005scape} and SHREC'19~\cite{MelziMRCRPWO19}, and an animal shape dataset SMAL~\cite{Zuffi_2017_CVPR}.
Next, in \cref{subsec:ImageMatching}, we further investigate the generality of \OurMethodName{} in the 2D domain with the PASCAL VOC dataset~\cite{everingham2010pascal}, in the context of \emph{keypoint matching} across natural images.

\subsection{3D Shape Matching}
\label{subsec:3DShapeMatching}

\myparagraphcpt{Datasets}
The FAUST and SCAPE datasets are widely used for evaluating human shape matching performance.
We follow prior works~\cite{ren2018continuous,roufosse2019unsupervised,Donati_2020_CVPR,eisenberger2020deep} and use the unaligned remeshed versions of the datasets, ensuring that shapes do not share identical mesh connectivity.
FAUST contains 100 shapes labeled with ground-truth correspondences,
and we use the same training/testing split as the prior works with 80/20 shapes, respectively.
SCAPE contains 71 labeled shapes
and is split into 51/20 shapes for training/testing.

Due to the nearly saturated performance on FAUST and SCAPE, we also test on SHREC'19~\cite{MelziMRCRPWO19}, a more challenging human shape matching dataset.
SHREC'19 contains 44 labeled shapes with the presence of a partial shape and 430 shape pairs in total.
SHREC'19 is used as a test set of generalizability, 
and we use the training sets of FAUST and SCAPE for network learning.

Finally, we also test on SMAL, a four-legged animal shape dataset
with five categories including cats, dogs, cows, horses, and hippos.
We use the first three categories as training data with 1,000 shapes per category.
The last two categories are only used as testing data with 100 shape pairs for non-isometric matching (\ie, one horse shape and one hippo shape in each testing pair).

\myparagraph{Implementation}
We use DiffusionNet~\cite{Sharp2020DiffusionNetDA} as the feature extractor $\mathcal{F}_\Theta$, which is a generic network for learning features on deformable shapes.
The local feature dimension $d$ is set to 128.
At training time, for the contrastive loss, we set $\tau = 0.07$ in \cref{eq:SoftPointCorrMatrix} and randomly sample 1,024 ground-truth correspondences.
We use
$s(\mathbf{x}, \mathbf{y}) = \mathbf{x} / \|\mathbf{x}\|_2 \cdot \mathbf{y} / \|\mathbf{y}\|_2$
to make the similarity measurement during training equivalent to the Euclidean distance metric used in proximity search at test time~\cite{grill2020bootstrap}.
We performed a simple parameter search to set $\lambda$ in \cref{eq:OurLoss}:
for the Dirichlet energy loss \cref{eq:DirichletEnergyLoss}, we set $\lambda=1$ on all the datasets;
for the spectral loss \cref{eq:SpectralFMapLoss}, we set $\lambda=0.1$ on FAUST and $\lambda=10$ on all the other datasets.
We use $k = 30$ eigenfunctions for the spectral loss, following~\cite{Donati_2020_CVPR}.

\myparagraph{Competitors}
In \cref{tab:faust_scape_shrec19_smal}, we perform comparisons to several recently proposed approaches for non-rigid shape matching.
The first category is axiomatic approaches, including 
BCICP~\cite{ren2018continuous}, 
ZoomOut~\cite{Melzi:2019:ZoomOut}, 
and Smooth Shells~\cite{eisenberger2020smooth}.
The second category is unsupervised learning approaches, including 
SURFMNet~\cite{roufosse2019unsupervised},
UnsupFMNet~\cite{Halimi_2019_CVPR},
NeuroMorph~\cite{Eisenberger_2021_CVPR},
and DeepShells~\cite{eisenberger2020deep}.
The third category is supervised learning approaches, including
FMNet~\cite{Litany_2017_ICCV},
3D-CODED~\cite{Groueix_2018_ECCV},
HSN~\cite{wiersma2020cnns},
ACSCNN~\cite{Li_2020_CVPR_ACSCNN},
DPFM\cite{attaiki2021dpfm},
and GeomFmaps~\cite{Donati_2020_CVPR}.
Note that for fair comparisons, we reproduced GeomFmaps with DiffusionNet, as done in~\cite{Sharp2020DiffusionNetDA}, which shows significantly better performance than the original KPConv-based GeomFmaps~\cite{Thomas_2019_ICCV,Donati_2020_CVPR}. DPFM is also based on DiffusionNet.
Post-processing techniques, such as ICP~\cite{ovsjanikov2012functional}, PMF~\cite{vestner2017product}, and ZO~\cite{Melzi:2019:ZoomOut}, may be used by the above approaches.

\begin{table}[ht]
   \centering
   \resizebox{\columnwidth}{!}{%
   \begin{tabular}{l@{\hspace{1em}}c@{\hspace{1em}}c@{\hspace{1em}}c@{\hspace{1em}}c@{\hspace{1em}}c@{\hspace{1em}}c}
                     \hline
                                         &               &               & \multicolumn{2}{c}{Train - Test}                                    &               \\
                     \cline{4-5}
                     Method              &       \tbf{F} &       \tbf{S} & \tbf{F} - \tbf{S} & \tbf{S} - \tbf{F} &                   \tbf{S19} & SMAL          \\
                     \hline                                                                                                                                      
                     BCICP               &           6.1 &          11.0 &                 - &                 - &                           - &             - \\
                     ZoomOut             &           6.1 &           7.5 &                 - &                 - &                           - &             - \\
                     SmoothShells        &           2.5 &           4.7 &                 - &                 - &                           - &             - \\
                     \hline                                                                                                                                      
                     SURFMNet            &          15.0 &          12.0 &              32.0 &              32.0 &                           - &             - \\
                    \ \ \tdf{\emph{+ICP}}&      \tdf{7.4}&      \tdf{6.1}&         \tdf{19.0}&         \tdf{23.0}&                      \tdf{-}&        \tdf{-}\\
                     UnsupFMNet          &          10.0 &          16.0 &              29.0 &              22.0 &                           - &             - \\
                    \ \ \tdf{\emph{+PMF}}&      \tdf{5.7}&     \tdf{10.0}&         \tdf{12.0}&          \tdf{9.3}&                      \tdf{-}&        \tdf{-}\\
                     NeuroMorph          &           8.5 &          29.9 &              28.5 &              18.2 &                           - &             - \\
                     DeepShells          &      \tul{1.7}&           2.5 &               5.4 &               2.7 &                        21.1 &          12.6 \\
                     \hline                                                                                                                                      
                     FMNet               &          11.0 &          17.0 &              30.0 &              33.0 &                           - &             - \\
                    \ \ \tdf{\emph{+PMF}}&      \tdf{5.9}&      \tdf{6.3}&         \tdf{11.0}&         \tdf{14.0}&                      \tdf{-}&        \tdf{-}\\
                     3D-CODED            &           2.5 &          31.0 &              31.0 &              33.0 &                           - &             - \\
                     HSN                 &           3.3 &           3.5 &              25.4 &              16.7 &                           - &             - \\
                     ACSCNN              &           2.7 &           3.2 &               8.4 &               6.0 &                           - &             - \\
                     DPFM                &           2.1 &           2.3 &          \tbf{2.7}&          \tul{2.5}&                         6.6 &           6.3 \\
                    \ \ \tdf{\emph{+ZO}} &      \tdf{1.9}&      \tdf{2.3}&          \tdf{2.4}&          \tdf{1.9}&                    \tdf{5.5}&      \tdf{5.5}\\
                     GeomFmaps           &           2.6 &           2.9 &          \tul{3.4}&               3.1 &                         8.5 &           6.0 \\
                    \ \ \tdf{\emph{+ZO}} &      \tdf{1.9}&      \tdf{2.6}&          \tdf{2.6}&          \tdf{1.9}&                    \tdf{7.9}&      \tdf{5.6}\\                                                                                                                  
                     CL                  &      \tbf{1.1}&      \tbf{1.9}&               6.1 &               3.7 &                        10.7 &          13.7 \\
                    \ \ \tdf{\emph{+ZO}} &      \tdf{1.9}&      \tdf{2.5}&          \tdf{2.8}&          \tdf{1.9}&                    \tdf{5.5}&      \tdf{5.0}\\
                     \OurMethodName{}-S  &      \tbf{1.1}&      \tul{2.2}&               3.9 &          \tul{2.5}&                    \tul{6.1}&      \tul{4.5}\\
                    \ \ \tdf{\emph{+ZO}} &      \tdf{1.9}&      \tdf{2.5}&          \tdf{2.6}&          \tdf{1.9}&                    \tdf{4.3}&      \tdf{5.3}\\
                     \OurMethodName{}-D  &      \tbf{1.1}&      \tbf{1.9}&               4.3 &          \tbf{2.2}&                    \tbf{5.4}&      \tbf{3.4}\\
                    \ \ \tdf{\emph{+ZO}} &      \tdf{1.9}&      \tdf{2.5}&          \tdf{3.1}&          \tdf{1.9}&                    \tdf{4.6}&      \tdf{4.9}\\
                    \hline
   \end{tabular}
   }
   \caption{Evaluation on the \tbf{F}AUST, \tbf{S}CAPE, \tbf{S}HREC'\tbf{19}, and SMAL datasets. The metric is mean geodesic error $\times 100$ on unit-area shapes. The results in \tdf{gray} are obtained by some specific post-processing techniques. The \textbf{best} and \underline{second best} results \emph{without} post-refinement are highlighted in each column.}
   \label{tab:faust_scape_shrec19_smal}
\end{table}

Contrastive learning (CL) is the straightforward baseline of our \OurMethodName{} framework, that is, we train the feature extractor with only the contrastive loss $\mathcal{L}_{\text{c}}$ ($\lambda = 0$ in \cref{eq:OurLoss}).
Though being simple, CL is not widely compared in the prior works on non-rigid shape matching, and presents a strong competitive baseline.
For our smoothness regularization approach, we denote CL with the Dirichlet energy loss as \OurMethodName{}-D, and CL with the spectral loss as \OurMethodName{}-S.
At test time, we compute correspondences between two input shapes by performing nearest-neighbor search between the learned local features for CL, \OurMethodName{}-D, and \OurMethodName{}-S.

\myparagraph{Results}
We use the evaluation metric introduced in~\cite{kim2011blended}, i.e., the mean geodesic error on unit-area shapes between the ground-truth and computed correspondences.
\cref{tab:faust_scape_shrec19_smal} shows comparisons on the FAUST, SCAPE, SHREC'19, and SMAL datasets.
For the sake of readability, we multiply the results by 100 and mark the results with post-refinement in \tdf{gray}.

On FAUST and SCAPE (\cref{tab:faust_scape_shrec19_smal}-left), our \OurMethodName{} shows competitive performance without post-refinement.
In the setting of training on FAUST and testing on SCAPE (\ie, \textbf{F} - \textbf{S}), DPFM and GeomFmaps perform better than \OurMethodName{}.
We ascribe this to the fact that the training set of FAUST contains a relatively small set of poses, which are different from those in SCAPE, and \OurMethodName{} is trained to be highly specialized on \textbf{F} with a very low testing error of 1.1, while DPFM and GeomFmaps have 2.1 and 2.6, respectively.
Interestingly, the feature matching based methods, \ie, CL, \OurMethodName{}-S, and \OurMethodName{}-D, achieve saturated performance in the settings of \textbf{F} and \textbf{S}, where post-refinement (\ie, +ZO) hurts the results, indicating that the results below 1.9 on \textbf{F} and 2.5 on \textbf{S} are nearly indistinguishable.
This motivates us to focus on the SHREC'19 and SMAL datasets for robustness and generalizability tests, as discussed below.
Nevertheless, \OurMethodName{} significantly improves CL in the cross dataset settings (\textbf{F} - \textbf{S} and \textbf{S} - \textbf{F}).

On SHREC'19 and SMAL (\cref{tab:faust_scape_shrec19_smal}-right), our \OurMethodName{}-D has the best matching performance.
We note that these experiments present a very challenging test, which involves non-isometric shape matching and evaluates generalization across datasets (\ie, training on FAUST + SCAPE, testing on SHREC'19) as well as across shape categories (\ie, no category overlap between the training and testing data of SMAL).
The experiments show that our \OurMethodName{} works well across various challenging testing scenarios and brings consistent improvement to CL.

\begin{figure}[t!]
	\centering
	\includegraphics[width=\linewidth]{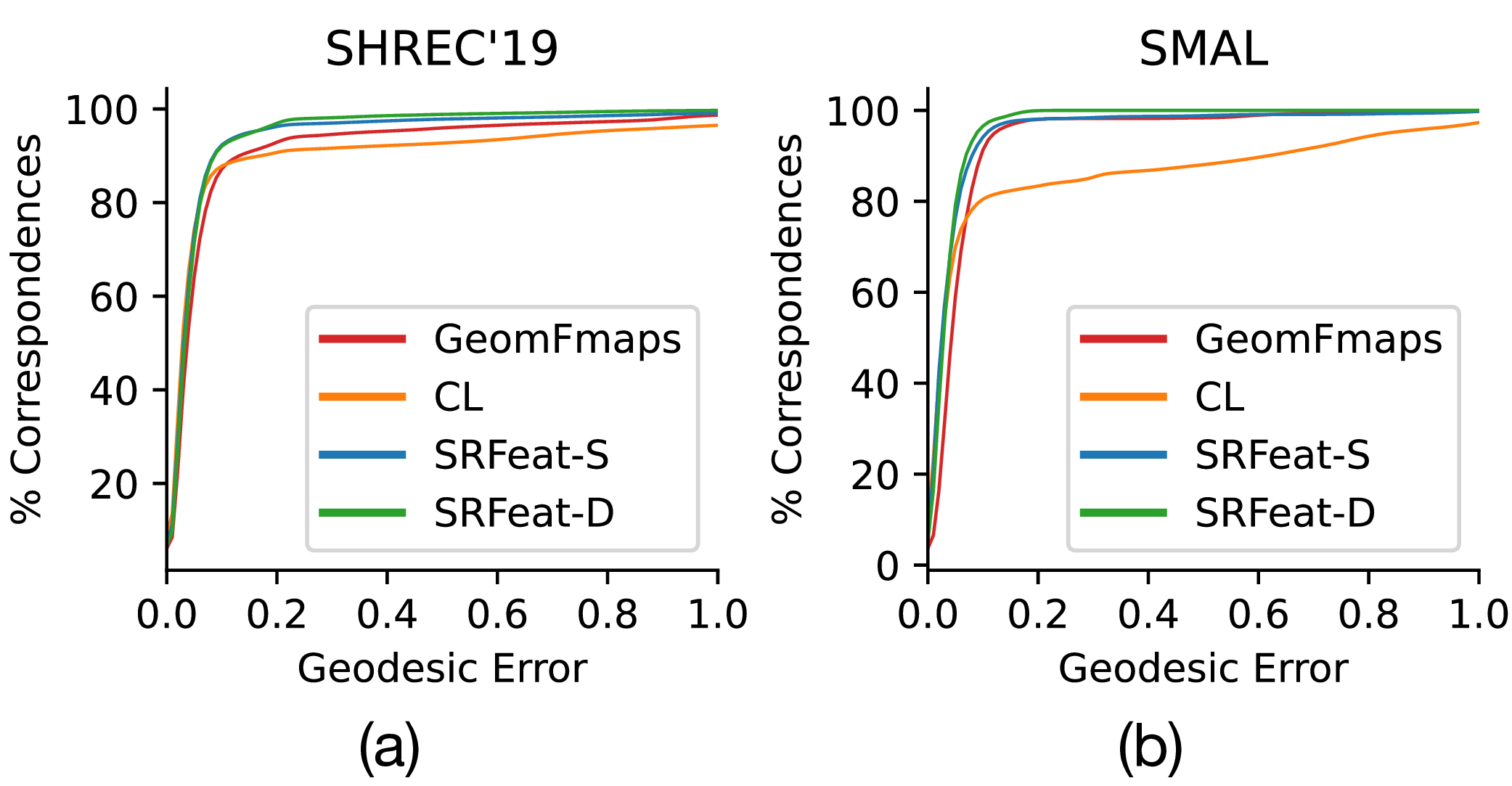}
    \caption{Correspondence quality evaluated on SHREC'19 and SMAL.}
    \label{fig:corr_geoerr}
\end{figure}

\begin{figure}[t!]
   \centering
   \includegraphics[width=\linewidth]{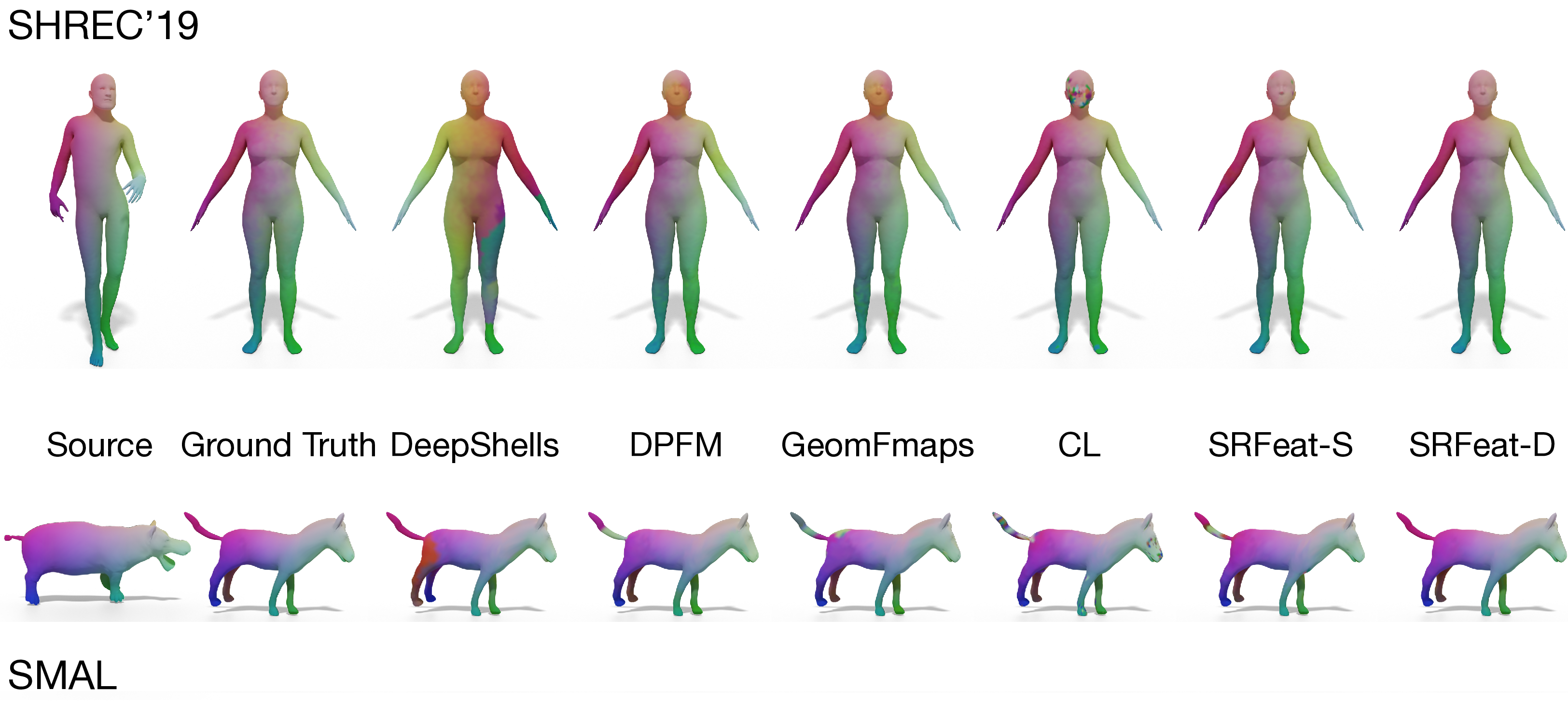}
   \caption{Qualitative results from the SHREC'19 and SMAL datasets \emph{without} using any post-refinement. Correspondence is visualized by color transfer.}
  \label{fig:qualitative_comparison}
\end{figure}

To get a better understanding of the improvement brought by smoothness regularization, in \cref{fig:corr_geoerr}, we plot cumulative curves showing the percentage of correspondences (y-axis) that have geodesic error smaller than a variable threshold (x-axis).
The plots show that \OurMethodName{}-S and \OurMethodName{}-D improve upon CL, and indicate, in particular, that using smoothness regularization in CL helps to significantly reduce \textit{correspondences with large errors} and thus improves the overall correspondence consistency and quality.
In \cref{fig:qualitative_comparison}, we show qualitative results \emph{without} performing any post-refinement, which can better reveal the original correspondence quality produced by each approach.
We observe that our \OurMethodName{} can produce more accurate correspondences while reducing both local and global artifacts.

\myparagraph{Ablation Study}
We perform a study \wrt the configurations of our approach.
First, we have shown the significant contribution of our smoothness regularization losses in \cref{tab:faust_scape_shrec19_smal} by comparing \OurMethodName{}-S and \OurMethodName{}-D with the baseline CL (\ie, setting $\lambda=0$ in the training loss \cref{eq:OurLoss}).

Next, we test different network architectures for feature extraction.
In the above experiments, we use DiffusionNet as the feature extractor, which needs connectivity for feature propagation.
In \cref{tab:smal_pointnet_sconv}, we test two recent architectures designed for learning on point sets: PointNet++~\cite{Qi:2017:PointNetPlusPlus} and SparseConv~\cite{Choy_2019_ICCV,Choy_2019_CVPR}.
We observe that \OurMethodName{}-S and \OurMethodName{}-D still improve CL noticeably, showing generality across architectures.
We find this result to be promising, as it shows that once trained on 3D data with mesh connectivity (used in the smoothness regularization losses), at test time our \OurMethodName{} can be applied to non-rigid point sets for local feature extraction and matching without the connectivity information.

\begin{figure}[t!]
	\centering
	\includegraphics[width=\linewidth]{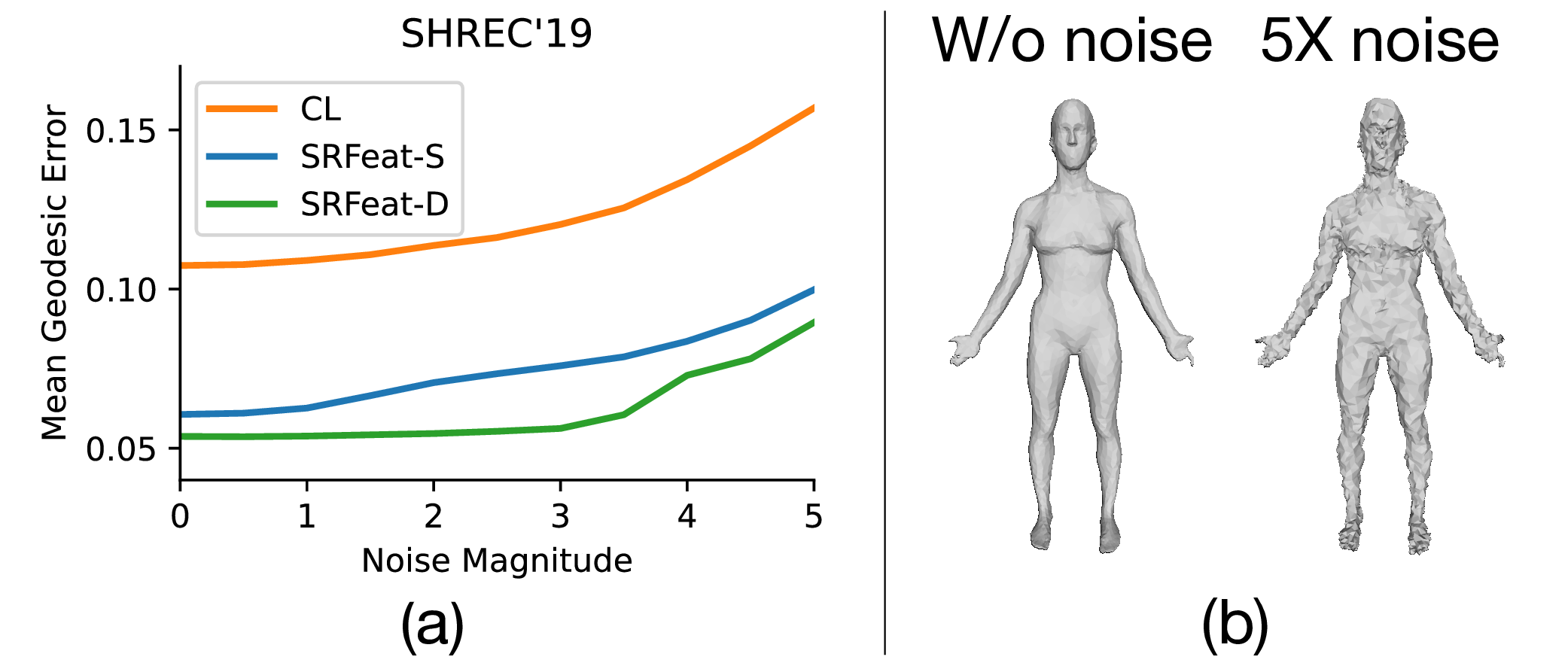}
	\caption{Test with varying noise magnitude on SHREC'19.}
	\label{fig:noise_test}
\end{figure}

Furthermore, we test the robustness of our \OurMethodName{} to input noise.
We add an increasing amount of Gaussian noise to point positions of shapes in the test set, as shown in \cref{fig:noise_test} (b), and we do not train or fine-tune the networks on each noise magnitude.
We plot the mean geodesic error \wrt the noise magnitude in \cref{fig:noise_test} (a).
We observe that \OurMethodName{}-S and \OurMethodName{}-D consistently improve CL across different noise levels and thus demonstrate the ability to handle moderate amounts of noise.

\begin{table}[t]
   \centering
   \resizebox{0.75\columnwidth}{!}{%
   \begin{tabular}{l@{\hspace{1em}}c@{\hspace{1em}}c@{\hspace{1em}}c}
      \hline
      Network        & CL                 & \OurMethodName{}-S & \OurMethodName{}-D \\
      \hline                                                                          
      PointNet++     &               6.2  &       \textbf{5.6} &               5.9  \\
      SparseConv     &               6.9  &               5.2  &       \textbf{3.9} \\
      \hline
   \end{tabular}
   }
   \caption{Shape matching performance of point-based feature extractors on SMAL (mean geodesic error $\times 100$).}
   \label{tab:smal_pointnet_sconv}
\end{table}

\begin{table*}[ht]
   \centering
   \resizebox{\textwidth}{!}{%
   \begin{tabular}{l|cccccccccccccccccccc|c}
       \hline
       Method                         &          Aero  &          Bike  &          Bird  &          Boat  &        Bottle  &           Bus  &           Car  &           Cat  &         Chair  &           Cow  &         Table  &           Dog  &         Horse  &        M-Bike  &        Person  &         Plant  &         Sheep  &          Sofa  &         Train  &           TV   &         Mean  \\  
       \hline
       GMN~\cite{Zanfir_2018_CVPR}    &          31.1  &          46.2  &          58.2  &          45.9  &          70.6  &          76.5  &          61.2  &          61.7  &          35.5  &          53.7  &          58.9  &          57.5  &          56.9  &          49.3  &          34.1  &          77.5  &          57.1  &          53.6  &          83.2  &          88.6  &         57.9  \\
       PCA-GM~\cite{wang2019learning} &          40.9  &          55.0  &  \textbf{65.8} &          47.9  &          76.9  &          77.9  &          63.5  &          67.4  &          33.7  &          66.5  &          63.6  &          61.3  &          58.9  &          62.8  &          44.9  &          77.5  &          67.4  &          57.5  &          86.7  &  \textbf{90.9} &         63.8  \\
       DGMC~\cite{FeyDGMC}            &          47.0  &          65.7  &          56.8  &          67.6  &          86.9  &          87.7  &          85.3  &          72.6  &  \textbf{42.9} &  \textbf{69.1} &          84.5  &          63.8  &  \textbf{78.1} &          55.6  &          58.4  &  \textbf{98.0} &          68.4  &          92.2  &  \textbf{94.5} &          85.5  &         73.0  \\
       \hline
       \OurMethodName{}-D             &          48.4  &  \textbf{69.4} &          57.3  &  \textbf{69.8} &  \textbf{88.2} &          86.0  &  \textbf{85.8} &  \textbf{73.3} &          42.1  &          67.7  &          93.2  &  \textbf{67.9} &          73.2  &          59.7  &          58.8  &          97.1  &          65.6  &          95.2  &          93.8  &          86.1  &         73.9  \\
       \OurMethodName{}-S             &  \textbf{49.2} &          69.1  &          57.6  &          68.3  &          87.7  &  \textbf{88.6} &          85.0  &          72.9  &          36.7  &          64.2  &  \textbf{95.1} &  \textbf{67.9} &          76.9  &  \textbf{65.0} &  \textbf{60.0} &          96.2  &  \textbf{68.6} &  \textbf{97.0} &          93.6  &          85.7  & \textbf{74.3} \\
       \hline
   \end{tabular}
   }
   \caption{Keypoint matching in natural images. Hits@1 (\%) on the PASCAL VOC dataset with Berkeley keypoint annotations.}
   \label{tab:image_kpt_matching}
\end{table*}

\subsection{Image Matching}
\label{subsec:ImageMatching}

To demonstrate the wide applicability of our smoothness regularization in other modern contrastive feature learning frameworks, we conduct experiments on an existing 2D image keypoint matching benchmark, as discussed in \cref{subsec:Background,subsec:GeneralizationTo2DMatching}.

\myparagraph{Dataset}
We follow~\cite{FeyDGMC} to test on the PASCAL VOC dataset~\cite{everingham2010pascal} with Berkeley keypoint annotations~\cite{bourdev2009poselets}.
The dataset has 6,953 and 1,671 natural images with annotated keypoints for training and testing, respectively.
Object instances in the images have varying scale, pose and illumination.
The number of keypoints in an object ranges from 1 to 19.

\myparagraph{Comparisons}
We perform comparisons with recent works GMN~\cite{Zanfir_2018_CVPR}, PCA-GM~\cite{wang2019learning}, and DGMC~\cite{FeyDGMC}.
DGMC is a recent deep graph matching network 
that performs node matching between two input graphs based on the node feature similarity, and  uses 
the contrastive loss $\mathcal{L}_{\text{c}}$ in \cref{eq:ContrastiveLoss} for training.

\myparagraph{Implementation}
Our implementation follows DGMC~\cite{FeyDGMC}.
Specifically, we use SplineCNN~\cite{Fey_2018_CVPR}, a graph neural network, as the feature extractor $\mathcal{F}_\Theta$.
We apply the Delaunay triangulation to the keypoints in each image for graph construction, which is used in both the graph neural network and our proposed smoothness regularization losses.
We augment the contrastive loss $\mathcal{L}_{\text{c}}$ with smoothness regularization, as done in \cref{eq:OurLoss}, where we set $\lambda=0.01$ for both the Dirichlet energy loss and the spectral loss.
We also use the same training and testing protocols as DGMC.

\begin{figure}[t!]
   \centering
   \includegraphics[width=\linewidth]{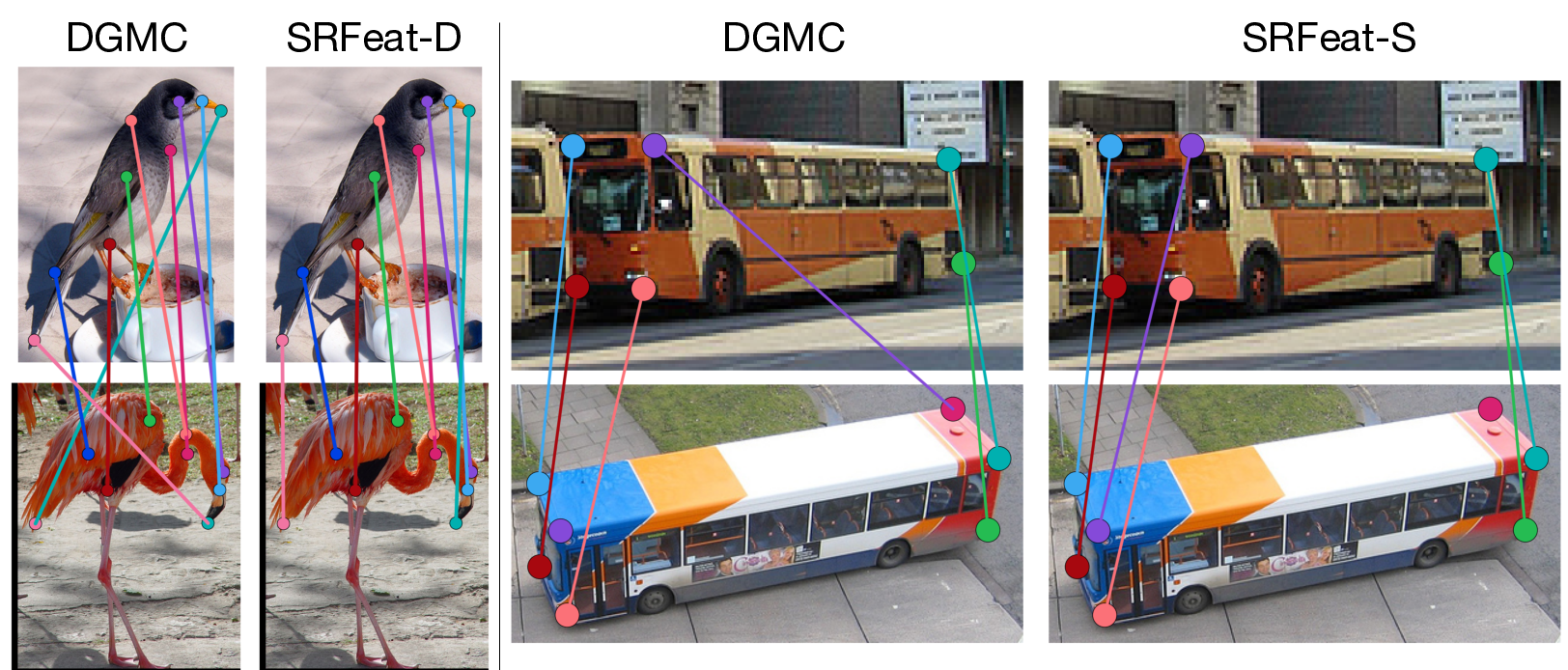}
   \caption{Qualitative results from the PASCAL VOC dataset. Ground-truth corresponding keypoints have the same color.}
  \label{fig:pascal_comparison}
\end{figure}

\myparagraph{Results}
\cref{tab:image_kpt_matching} presents the performance comparisons on PASCAL VOC.
The evaluation metric is Hits@1, measuring the proportion of correctly matched keypoints ranked in the top-1.
\cref{fig:pascal_comparison} further shows qualitative comparisons.
We observe that our smoothness regularization brings noticeable improvement over DGMC, which uses contrastive learning only, and \OurMethodName{}-S achieves the best performance on this benchmark.
We ascribe the slightly better performance of \OurMethodName{}-S over \OurMethodName{}-D partly to the fact that image objects have very sparse ($\leq 19$) keypoints and may undergo significant geometric distortion and occlusion (\cref{fig:pascal_comparison}).
Thus enforcing approximate smoothness and global structure consistency by the spectral loss in a reduced basis can be more beneficial.
In contrast, 3D shape matching in \cref{subsec:3DShapeMatching} is a different application scenario, where dense ($\sim5,000$) point correspondence is tested.
The Dirichlet energy loss encourages the features to vary smoothly on 3D surfaces to better capture the underlying geometry, resulting in more precise dense matching.
Nevertheless, the result in \cref{tab:image_kpt_matching} strongly indicates the general applicability of our smoothness-regularized feature learning framework.

\section{Conclusion, Limitations \& Future Work}
\label{sec:Conclusion}

In this work, we have presented \OurMethodName{}, a generic learning-based framework that combines the local accuracy of contrastive learning with the global consistency of geometric approaches, for robust non-rigid shape correspondence.
Through extensive experiments, we show that \OurMethodName{} produces discriminative local features that can be robustly matched, resulting in more accurate correspondences.
We demonstrate the effectiveness and generality of \OurMethodName{} on a suite of benchmarks including 3D non-rigid shape matching and 2D image keypoint matching.

One limitation of our approach is that to compute smoothness regularization we assume the input shapes to be represented as triangle meshes during training, and it might be worth extending our approach to point clouds.
For partially overlapped shapes, unlike \cite{attaiki2021dpfm,huang2020predator}, \OurMethodName{} does not have a specific component for predicting overlapping masks used in matching, which might be worth further investigation.
Unsupervised learning is another interesting direction to explore.
To eschew ground-truth annotations, one possible approach is, like~\cite{xie2020pointcontrast}, to create synthetic shape pairs via data augmentations. A key challenge would be to systematically study various augmentation strategies for deformable shapes that can lead to informative local feature learning.
Finally, further improvements to the matching quality of \OurMethodName{} can be brought, \eg, by incorporating a learnable correspondence filtering component~\cite{choy2020deep,bai2021pointdsc}, in future work.

\section*{Acknowledgements}
The authors thank the anonymous reviewers for their valuable comments and suggestions. Parts of this work were supported by the ERC Starting Grant No. 758800 (EXPROTEA) and the ANR AI Chair AIGRETTE.

{\small
\bibliographystyle{ieee_fullname}
\bibliography{egbib}
}

\newpage
\appendix
In this supplementary material, we first include more implementation details in \cref{sec:ImplementationDetailsSupp}.
Next, we perform further investigation of our spectral loss in \cref{sec:SpectralLossSupp}.
Finally, we present more quantitative and qualitative results in \cref{sec:MoreResultsSupp}.

\section{Implementation Details}
\label{sec:ImplementationDetailsSupp}

\subsection{3D Shape Matching}
We use DiffusionNet~\cite{Sharp2020DiffusionNetDA} as the feature extraction backbone, and its implementation is based on the publicly available codebase\footnote{\url{https://github.com/nmwsharp/diffusion-net}} released by its authors.
The network is composed of four DiffusionNet blocks of width 128.
The network takes 3D point positions (\ie, xyz) as input signals and outputs 128-dimensional point-wise features.
We set the batch size to 1 and use the ADAM optimizer~\cite{KingmaB14:Adam} with an initial learning rate of 0.001.
We use servers equipped with NVIDIA A100 and GeForce GTX 1080 GPUs for network training.

In \cref{tab:lambda_sweep_supp}, we show the matching performance \wrt{} $\lambda$ in Eq.~(4) on the SHREC'19 and SMAL datasets,
resulting in the choice $\lambda=10$ for the spectral loss Eq.~(9) and $\lambda=1$ for the Dirichlet energy loss Eq.~(7).

In \cref{tab:tab_runtime_supp}, we show the runtime of shape matching on the SMAL dataset.
The statistics were collected on a server with AMD EPYC 7302 CPU, 512GB RAM, and NVIDIA A100 GPU.
The columns \emph{Feature}, \emph{FMap}, and \emph{PMap} represent the runtime of feature extraction by DiffusionNet, functional map computation, and point-wise map computation with k-d tree, respectively.
We reiterate that the feature matching based methods, \ie, CL, \OurMethodName{}-S, and \OurMethodName{}-D, do \emph{not} require functional map computation at test time.
Note that the distribution of high dimensional features can affect the nearest neighbor search performance of k-d trees used in the point-wise map computation.
Nevertheless, \OurMethodName{}-D has the best runtime performance.

\subsection{Image Matching}
We incorporate our proposed smoothness regularization in DGMC~\cite{FeyDGMC} for the 2D image keypoint matching task.
Specifically, we build upon the publicly available codebase\footnote{\url{https://github.com/rusty1s/deep-graph-matching-consensus}} of DGMC, which is trained with only a contrastive loss, as mentioned in Sec.~3.1 of the main text.
In the pre-processing stage, each image is forwarded through a pre-trained VGG16 network, and features of the annotated image keypoints are then extracted on the \texttt{relu4\_2} and \texttt{relu5\_1} feature maps through bilinear interpolation and concatenated together.
DGMC adopts SplineCNN, a graph neural network, to extract 256-dimensional node-wise features for matching.
Delaunay triangulation is used to construct a graph for the keypoints in each image.
To incorporate our smoothness regularization in the training loss, we reuse the triangulation result for the Laplacian matrix construction, which is required in Eq.~(6) and (8) of the main text.
We set the batch size to 512 and use the ADAM optimizer with a learning rate of 0.001.
The network is trained for 15 epochs.

\begin{table}[t]
   \centering
   \resizebox{\columnwidth}{!}{%
   \begin{tabular}{l|cc|cc}
      \hline
                       & \multicolumn{2}{c|}{SHREC'19}           & \multicolumn{2}{c}{SMAL}                \\
                       & \OurMethodName{}-S & \OurMethodName{}-D & \OurMethodName{}-S & \OurMethodName{}-D \\
       \hline
       $\lambda=0.1$   &               11.2 &                7.3 &               14.4 &                4.8 \\
       $\lambda=1$     &               10.4 &           \tbf{5.4}&                8.6 &           \tbf{3.4}\\
       $\lambda=10$    &           \tbf{6.1}&                7.2 &           \tbf{4.5}&                5.0 \\
       $\lambda=100$   &               11.1 &               37.2 &                6.6 &               10.7 \\
       \hline
   \end{tabular}
   }
   \caption{Matching performance \wrt{} $\lambda$ on the SHREC'19 and SMAL datasets (mean geodesic error $\times 100$ on unit-area shapes).}
   \label{tab:lambda_sweep_supp}
\end{table}

\begin{table}[t]
   \centering
   \resizebox{0.9\columnwidth}{!}{%
   \begin{tabular}{l|ccc|c}
      \hline
                                       &             Feature  &             FMap  &             PMap  &       Total \\
      \hline
      GeomFmaps                        &              0.0226  &           0.0437  &           0.0215  &      0.0878 \\
      CL                               &              0.0227  &                -  &           0.0794  &      0.1021 \\
       SRFeat-S                         &              0.0227  &                -  &           0.0773  &      0.1000 \\
       SRFeat-D                         &              0.0225  &                -  &           0.0419  & \tbf{0.0644}\\
      \hline
   \end{tabular}
   }
   \caption{Runtime (s) per shape pair averaged on SMAL.}
   \label{tab:tab_runtime_supp}
\end{table}

\section{Spectral Loss}
\label{sec:SpectralLossSupp}

In Eq.~(8) of the main text, we propose to compute a functional map directly from a learned soft point-wise map within the network.
In this section, we perform further investigation of Eq.~(8) and compare it with the FMReg layer proposed in GeomFmaps~\cite{Donati_2020_CVPR}.

GeomFmaps computes a functional map by treating learned features as probe functions and solving an energy minimization problem in the spectral domain (see Sec.~4.4 of~\cite{Donati_2020_CVPR}), which is referred as the FMReg layer.
This layer, however, needs to solve multiple linear systems within the network, and requires differentiating through the matrix inverse, and thus can be computationally demanding and numerically unstable during training as observed in existing literature~\cite{roufosse2019unsupervised,Donati_2020_CVPR}.

We also compared our proposal based on the definition given in Eq.~(8) of the main text, with the FMReg layer introduced in ~\cite{Donati_2020_CVPR}.
For this, we directly replace the FMReg layer in GeomFmaps with our Eq.~(8) to compute the functional map $\mathbf{C}$, which is compared to the ground-truth $\mathbf{C}_{\text{gt}}$ as the training loss.
The rest of the GeomFmaps network is kept the same.

We remark that GeomFmaps w/ Eq.~(8) studied in this additional experiment is a  variant of the functional map approaches for shape correspondence, which is \emph{different} from the feature matching based methods \OurMethodName{}-S and \OurMethodName{}-D in our main text.
Specifically, GeomFmaps w/ Eq.~(8) does not use any contrastive learning losses, and requires the Laplacian basis computation and the functional map estimation at test time.

In \cref{tab:spectral_loss_comparison_supp}, we show the matching performance on the SHREC'19 and SMAL datasets.
We observe that Eq.~(8) improves GeomFmaps on SHREC'19 and has comparable performance with the FMReg layer on SMAL.
Note that the performance of our \OurMethodName{}-S and \OurMethodName{}-D has been reported in Tab.~1 of the main text.
We further show the runtime statistics in \cref{tab:spectral_loss_runtime_comparison_supp} and observe that Eq.~(8) significantly speeds up the functional map computation by two orders of magnitude (from 0.0437s to 0.0004s) and reduces the overall runtime by more than a \emph{half}.

\begin{table}[t]
   \centering
   \resizebox{0.7\columnwidth}{!}{%
   \begin{tabular}{l|cc}
      \hline
                                       &   SHREC'19 &       SMAL \\
      \hline
      w/ FMReg~\cite{Donati_2020_CVPR} &        8.5 &  \tbf{6.0 }\\
      w/ Eq. (8)                       &   \tbf{5.8}&       6.1  \\
      \hline
   \end{tabular}
   }
   \caption{Matching performance of GeomFmaps~\cite{Donati_2020_CVPR} with different functional map computation schemes (mean geodesic error $\times 100$ on unit-area shapes).}
   \label{tab:spectral_loss_comparison_supp}
\end{table}

\begin{table}[t]
   \centering
   \resizebox{0.9\columnwidth}{!}{%
   \begin{tabular}{l|ccc|c}
      \hline
                                       &             Feature  &             FMap  &             PMap  &       Total \\
      \hline
      w/ FMReg~\cite{Donati_2020_CVPR} &              0.0226  &           0.0437  &           0.0215  &      0.0878 \\
       w/ Eq. (8)                       &              0.0226  &           0.0004  &           0.0193  & \tbf{0.0423}\\
      \hline
   \end{tabular}
   }
   \caption{Runtime (s) of GeomFmaps~\cite{Donati_2020_CVPR} with different functional map computation schemes on SMAL.}
   \label{tab:spectral_loss_runtime_comparison_supp}
\end{table}

\section{More Results}
\label{sec:MoreResultsSupp}

In \cref{tab:srfeat-s-d_shrec19_smal}, we show the performance of \OurMethodName{}-S-D, which combines CL with the spectral and Dirichlet energy losses. We performed a hyperparameter search to set weights for the spectral and Dirichlet energy losses, resulting in $(0.1, 1)$ for SHREC'19, and $(0.1, 0.1)$ for SMAL. Observe that this variant slightly outperforms \OurMethodName{}-S but is comparable to \OurMethodName{}-D, indicating that the Dirichlet energy regularization is sufficient for contrastive learning on 3D shapes. \OurMethodName{}-S-D requires more hyperparameter tuning, which may be undesirable in practice.

\begin{table}[h]
   \centering
   \resizebox{0.65\columnwidth}{!}{%
   \begin{tabular}{lcc}
   \hline
   Method                      & SHREC'19          & SMAL          \\
   \hline
   \OurMethodName{}-S          &              6.1  &          4.5  \\
   \OurMethodName{}-D          &              5.4  &          3.4  \\
   \hline
   \OurMethodName{}-S-D        &              5.3  &          3.5  \\
   \hline
   \end{tabular}
   }
   \caption{Matching performance of \OurMethodName{}-S-D (mean geodesic error $\times 100$ on unit-area shapes).}
   \label{tab:srfeat-s-d_shrec19_smal}
\end{table}

\begin{figure*}[!ht]
   \centering
   \includegraphics[width=\linewidth]{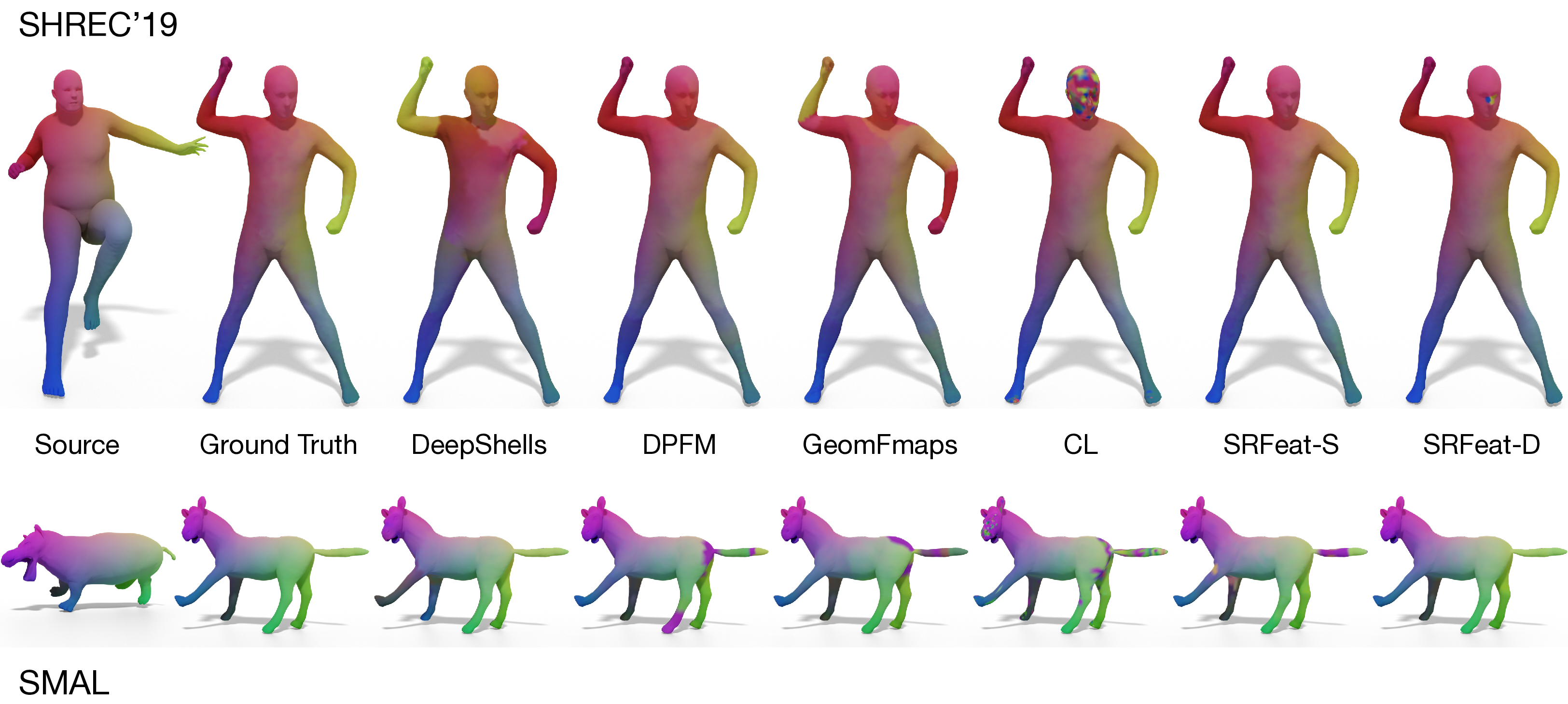}
   \caption{More qualitative results from the SHREC'19 and SMAL datasets \emph{without} using any post-refinement. Correspondence is visualized by color transfer.}
  \label{fig:qualitative_comparison_supp}
\end{figure*}

In \cref{fig:qualitative_comparison_supp}, we present more qualitative results of non-rigid shape matching on the SHREC'19 and SMAL datasets.
We note that the matching results are obtained \emph{without} performing any post-refinement, which shows the original matching quality of each method. 
While \OurMethodName{} may not be completely free from correspondence outliers, the results show that our smoothness regularization brings noticeable improvement to the matching quality of CL.

\begin{figure*}[!ht]
   \centering
   \includegraphics[width=\linewidth]{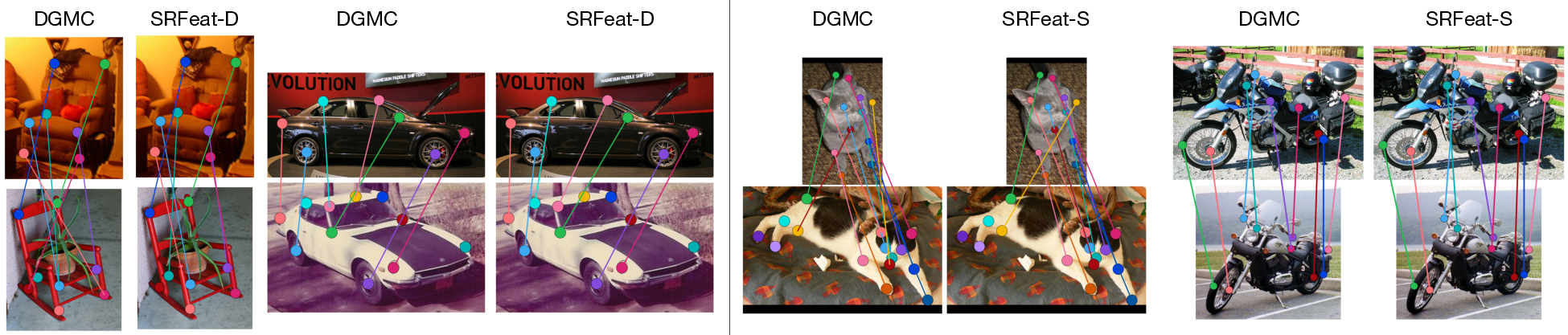}
   \caption{More qualitative results from the PASCAL VOC dataset. Ground-truth corresponding keypoints have the same color.}
  \label{fig:pascal_comparison_supp}
\end{figure*}

In \cref{fig:pascal_comparison_supp}, we also present more qualitative results of the 2D image keypoint matching task on the PASCAL VOC dataset, demonstrating the improvement of \OurMethodName{} over DGMC.

\end{document}